\documentclass[sigconf]{acmart}

\usepackage{booktabs} % For formal tables
\usepackage{amsmath}
\usepackage{xcolor,tikz,colortbl}
\usepackage{listings} 

\setcopyright{rightsretained}
\begin{document}

\title{LibOPT: An Open-Source Platform for Fast Prototyping Soft Optimization Techniques}

\author{J.P. Papa}
\affiliation{%
  \institution{S\~ao Paulo State University}
  %\streetaddress{P.O. Box 1212}
  \city{Bauru} 
  \state{S\~ao Paulo, Brazil} 
  \postcode{17033-360}
}
\email{papa@fc.unesp.br}

\author{G.H. Rosa}
\affiliation{%
  \institution{S\~ao Paulo State University}
%  \streetaddress{P.O. Box 1212}
 \city{Bauru} 
  \state{S\~ao Paulo, Brazil} 
  \postcode{17033-360}}
\email{gustavo.rosa@fc.unesp.br}

\author{D. Rodrigues}
\affiliation{%
  \institution{Federal University of S\~ao Carlos}
%  \streetaddress{1 Th{\o}rv{\"a}ld Circle}
  \city{S\~ao Carlos} 
  \state{S\~ao Paulo, Brazil}
  \postcode{13565-905}}
\email{douglasrodrigues.dr@gmail.com}

\author{X.-S. Yang}
\affiliation{%
  \institution{Middlesex University}
  \city{Hendon}
  \state{London, UK} 
  \postcode{NW4 4BT}}
\email{x.yang@mdx.ac.uk}

% The default list of authors is too long for headers}
\renewcommand{\shortauthors}{Papa et. al.}

\begin{abstract}
Optimization techniques play an important role in several scientific and real-world applications, thus becoming of great interest for the community. As a consequence, a number of open-source libraries are available in the literature, which ends up fostering the research and development of new techniques and applications. In this work, we present a new library for the implementation and fast prototyping of nature-inspired techniques called LibOPT. Currently, the library implements $15$ techniques and $112$ benchmarking functions, as well as it also supports $11$ hypercomplex-based optimization approaches, which makes it one of the first of its kind. We showed how one can easily use and also implement new techniques in LibOPT under the C paradigm. Examples are provided with samples of source-code using benchmarking functions.
\end{abstract}



% Keywords

% Paper
\maketitle

\section{Introduction}
\label{s.introduction}

Soft computing concerns designing techniques that can provide inexact though reasonable solutions to a given problem in a feasible amount of time. A branch of such approaches is represented by the so-called ``bio-inspired"\ techniques, which aim at solving optimization problems by means of nature-driven heuristics~\cite{Yang:08}. Also known as ``metaheuristics", such techniques play an important role in both scientific and commercial communities.

The research area devoted to the study and development of optimization techniques based on nature and living beings has grown widely in the last decades. Approaches based on bird flocks~\cite{Kennedy:01}, bats~\cite{YangBA:12}, fireflies~\cite{YangFFA:10}, ants~\cite{Dorigo:04}, and bees~\cite{KarabogaJGO:07} are some successful examples of techniques that have been used in a number of optimization-oriented applications, just to name a few.

As a consequence, one may refer to several optimization libraries based on metaheuristics in the literature\footnote{\url{https://pypi.python.org/pypi/metaheuristic-algorithms-python/0.1.6}}\footnote{\url{http://opt4j.sourceforge.net}}\footnote{\url{https://github.com/yasserglez/metaheuristics}}\footnote{\url{http://home.gna.org/momh/}}\footnote{\url{https://projects.coin-or.org/metslib}}\footnote{\url{http://dev.heuristiclab.com/trac.fcgi/wiki/Download}}~\cite{EgeaBMC:14,Beukelaer:16}, some of them being implemented in Java, C++, C\#, Python, R and Matlab. There are a plenty of techniques implemented in such libraries, which range from  population- and evolutionary-based optimization techniques to multi-objective algorithms. Some libraries even consider machine learning techniques, thus making available a bunch of tools to be further used by users and readers.

Although one may refer to a number of libraries, most of them have their bottlenecks, such as do not being self-contained, i.e., they need other libraries to be installed together, and some of them comprise a number of techniques other than optimization ones, which may take longer to the user to get used to it. In this work, we present the LibOPT, which is an open-source library implemented in C language for the development and use of metaheuristic-based optimization techniques. The library is self-contained, which means it does not require additional packages, it contains only optimization techniques, and it can be easily integrated with other tools. Currently, LibOPT implements $15$ techniques, $112$ benchmarking functions, and it also contains $11$ hypercomplex-oriented optimization techniques based on quaternions and octonions. Actually, the library allows the user to implement any number of hypercomplex dimensions. As far as we are concerned, LibOPT is the only one that can group all these features, mainly with respect to the hypercomplex-based search spaces.

\begin{sloppypar}
LibOPT has been used in a number of scientific works that comprise feature selection~\cite{RamosIEEETPD:12,RodriguesESA:14,RodriguesRASIEC:15,RamosIEEETPG:16}, deep learning fine tuning~\cite{PapaGECCO:15,PapaASC:16,PapaJoCS:15,RosaCIARP:15,RosaANNPR:16}, pattern recognition~\cite{PapaPRL:17}, hyperheuristics~\cite{PapaJSTARS:16}, and quaternion-based optimization~\cite{PapaANNPR:16}, among others. The library is built upon the concept of fast prototyping, since all techniques share the very same underlying idea, which means that it is quite easy to implement additional techniques. Also, if one desires to use the library instead of going deeper into the metaheuristic designing, the only thing to be implemented is the function to be minimized.
\end{sloppypar}

In this paper, we present the main functionalities concerning LibOPT, as well as how one can simply use it in two different ways: (i) just for application purposes, i.e., we want to minimize some function, or for (ii) development purposes, i.e., we want to implement a new metaheuristic optimization technique in the library. The paper provides a comprehensive but quite simple viewpoint of the main data structures used in the library, and how one can install and configure the package for further usage.

The remainder of the paper is organized as follows. Section~\ref{s.libopt} presents the LibOPT and its main functionalities, and Section~\ref{s.howto} addresses a ``how to"-like approach, i.e., how one can use the library to minimize her/his own function, as well as how to add a new technique. Finally, Section~\ref{s.conclusions} states conclusions and future works.

\section{Library Organization}
\label{s.libopt}

In this section, we present the main tools implemented in LibOPT, as well as how to install the library and design your own function.

\subsection{General Tools}
\label{ss.metaheuristics}

LibOPT is freely available at GitHub\footnote{\url{https://github.com/jppbsi/LibOPT}}, where a home-page presents all techniques\footnote{\url{https://github.com/jppbsi/LibOPT/wiki}} and benchmarking functions currently available in the library. To date, LibOPT comprises $15$ techniques, as follows:

\begin{itemize}
	\item Particle Swarm Optimization~\cite{Kennedy:01};
	\item Particle Swarm Optimization with Adaptive Inertia Weight~\cite{NickabadiASC:11};
	\item Bat Algorithm~\cite{YangBA:12};
	\item Flower Pollination Algorithm~\cite{YangFPA:14};
	\item Firefly Algorithm~\cite{YangFFA:10};
	\item Cuckoo Search~\cite{YangIJMMNO:10};
	\item Genetic Programming~\cite{Koza:92};
	\item Black Hole Algorithm~\cite{HatamlouIS:13};
	\item Migrating Birds Optimization~\cite{Duman:12};
	\item Geometric Semantic Genetic Programming~\cite{MoraglioICPPSN:12};
	\item Artificial Bee Colony~\cite{KarabogaJGO:07};
	\item Water Cycle Algorithm~\cite{EskandarCS:12};
	\item Harmony Search~\cite{Geem:09};
	\item Improved Harmony Search~\cite{MahdaviAMC:07}; and
	\item Parameter-setting-free Harmony Search~\cite{GeemAMC:10}.
\end{itemize}
As one can observe, the library comprises a broad variety of techniques, such as population- and evolutionary-based, phenomenon-mimicking, and nature-inspired.

In regard to hypercomplex-based techniques, LibOPT implements the following approaches concerning quaternion- and octonion-based representations:

\begin{itemize}
\item Particle Swarm Optimization;
\item Particle Swarm Optimization with Adaptive Inertia Weight;
\item Bat Algorithm~\cite{FisterIEEECEC:15};
\item Flower Pollination Algorithm;
\item Firefly Algorithm~\cite{FisterESA:13};
\item Cuckoo Search;
\item Black Hole Algorithm;
\item Artificial Bee Colony;
\item Harmony Search~\cite{PapaANNPR:16};
\item Improved Harmony Search~\cite{PapaANNPR:16}; and
\item Parameter-setting-free Harmony Search.
\end{itemize}
Notice that most of the above techniques in their hypercomplex representation are not even published yet. Additionally, LibOPT implements $112$ benchmarking functions\footnote{\url{https://github.com/jppbsi/LibOPT/wiki/Benchmarking-functions}}~\cite{Jamil13}, which are not displayed here for the sake of space.

\subsection{Installation}
\label{ss.installation}

The library was implemented and tested to work under Unix- and MacOS-based operational systems, and it can be quickly installed by executing the \verb|make| command right after decompressing the file. On MacOS, if one faces any problem, you should try using GNU/gcc compiler\footnote{\url{https://github.com/jppbsi/LibOPT/wiki/Installation}}.

\subsection{Data Structures}
\label{ss.structures}

Apart from other directories, LibOPT contains two main folders, say that \verb|LibOPT\include| and \verb|LibOPT\src|, being the first one in charge of the header files, and the latter responsible for the source files and main implementation.

In order to allow a fast prototyping, the library was created with one main structure in mind called \verb|Agent|, which has the following implementation in its simplest version:

\begin{verbatim}
typedef struct Agent_{
    /* common definitions */
    int n; /* number of decision variables */
    double *x; /* position */
    double fit; /* fitness value */
    double **t; /* tensor */
}Agent;
\end{verbatim} 
The above implementation comprises all common information shared by the techniques implemented to date, which means all techniques available in the library must set that parameters, being $n$ the number of decision variables to be optimized, and $\textbf{x}$ an array that encodes the current position of the agent when working under standard search spaces. Further, variable \verb|fit| stores the fitness value, and $\textbf{t}$ stands for a matrix-like structure that is used to implement the hypercomplex-based versions of the na\"ive techniques, and it works similarly to $\textbf{x}$, but in another search space representation.

Another main structure models the whole search space, which includes additional information concerning the optimization problem other than the agents, as follows:

\begin{verbatim}
typedef struct SearchSpace_{
    /* common definitions */
    int m; /* number of agents (solutions) */
    int n; /* number of decision variables */
    int iterations; /* number of iterations */
    Agent **a; /* array of pointers to agents */
    double *LB; /* lower boundaries */
    double *UB; /* upper boundaries */
    double *g; /* global best agent */
    double **t_g; /* global best tensor */
    int best; /* index of the best agent */
    double gfit; /* global best fitness */
    int is_integer_opt; /* integer-valued problem? */
}SearchSpace;
\end{verbatim}
Notice the library contains a quite detailed explanation about every attribute information in order to avoid possible misunderstandings, thus leading the user to the maximum advantages of LibOPT.

The main purpose of \verb|SearchSpace| structure is to encode crucial information about the optimization problem, such as the number of agents (solutions) $m$, the lower and upper boundaries of each decision variable, the global best agent and the global best fitness, among others. As the reader may have noticed,  \verb|SearchSpace| also includes the number of decision variables (dimensionality of the search space), although \verb|Agent| structure contains the very same information. The reason for that is related to the fitness function, as shall be explained later, which has as the main input parameter an \verb|Agent| structure instead of the whole search space. Thus, we need \verb|Agent| to be self-contained.

\begin{sloppypar}
Both \verb|Agent| and \verb|SearchSpace| structures are defined in \verb|LibOPT/include/common.h|, as well as other common structures and functions, and their implementations can be found in \verb|LibOPT/src/common.c|. In order to facilitate the allocation and deallocation of every structure in LibOPT, the library comprises constructors and destructors, similarly to an implementation in C++. As an example, we have the constructor \verb|Agent *CreateAgent(int n, int opt_id)|, which has the number of dimensions (decision variables) and the identification of the metaheuristic technique that is going to be considered. For instance, one can create an agent with $10$ dimensions related to the Particle Swarm Optimization (PSO) technique as follows: \verb|Agent A = CreateAgent(10, _PSO_)|, where \verb|_PSO_| is the directive concerning PSO. More detailed information about that directives are given further. The deallocation of that agent can be easily implemented using the following command: \verb|DestroyAgent(&A, _PSO_)|.
\end{sloppypar}

\subsection{Model Files}
\label{ss.model_files}

As aforementioned, although most of techniques have something in common (e.g., number of decision variables, current position and maybe velocity), they may also differ in the number of parameters. Such circumstance led us to design a model file-based implementation, which means all parameter setting up required for a given optimization technique must be provided in a single text file, hereinafter called ``model file".

For the sake of explanation, let us consider the model file of Particle Swarm Optimization\footnote{Detailed information concerning the model files of the techniques implemented in LibOPT can be found at \url{https://github.com/jppbsi/LibOPT/wiki/Model-files}.}. Roughly speaking, the user must input all information required by that technique, as follows:

\begin{verbatim}
10 2 100 #<n_particles> <dimension> <max_iterations>
1.7 1.7 #<c1> <c2>
0.7 0.0 0.0 #<w> <w_min> <w_max>
-5.12 5.12 #<LB> <UB> x[0]
-5.12 5.12 #<LB> <UB> x[1]
\end{verbatim}

The first line contains three integers: number of agents (particles), number of decision variables (dimension) and number of iterations. Notice everything right after the caracter \# is considering a comment, thus not taking into account by the parser. The next two lines configure PSO parameters $c_1$ and $c_2$, and the inertia weight $w$. Since LibOPT implements the na\"ive PSO, it does not employ adaptive inertia weight (they are used only for Particle Swarm Optimization with Adaptive Inertia Weight). Therefore, there is no need to set $w_{min}$ and $w_{max}$. The last two lines aim at setting up the range of each decision variable. Since we have two dimensions in the example, each line stands for one variable, say $x[0]$ and later $x[1]$. In the above example, we have a problem with $10$ particles, $2$ decision variables and $100$ iterations for convergence. Also, we used $c_1=c_2=1.7$, $w=0.7$, and $x[i]\in[-5.12,5.12]$, $i\in\{0,1\}$.

\section{Using LibOPT}
\label{s.howto}

In this section, we present one toy example concerning using LibOPT to optimize your own function, and another one discussing how to add a new technique in there.

\subsection{Function Optimization}
\label{ss.function_optimization}

LibOPT works with the concept of ``function minimization", which means you need to take that into account when trying to ``maximize"\ some function. Suppose we want to minimize the following 2D function:

\begin{equation}
	f(x, y) = x^2 + y^2 + 1,
\end{equation}
\begin{sloppypar}
\noindent where $x, y \in[-10,10]$ and $x, y\in\Re$. Note that for simplicity reasons, we will be using $x$ as $x_0$ and $y$ as $x_1$. Since all functions are implemented in both \verb|LibOPT/include/function.h| (header) and  \verb|LibOPT/src/function.c| directories, one must add the function's signature in the first file, and the function's implementation in the second one.
\end{sloppypar}

In \verb|LibOPT/include/function.h|, the following line of code must be added: \verb|double MyFunction(Agent *a, va_list arg);|. With respect to the file \verb|LibOPT/src/function.c|, one should implement the function as follows:

\begin{lstlisting}
double MyFunction(Agent *a, va_list arg){
    double output;
    
    if(!a){
        fprintf(stderr,"\nAgent not allocated @MyFunction.");
        return DBL_MAX;
    }
    if(a->n < 1){
        fprintf(stderr,"\nInvalid number of decision variables @MyFunction. It must be equal or greater than one.\n");
        return DBL_MAX;
    }
    output = pow(a->x[0], 2) + pow(a->x[1], 2) + 7; /* Equation (1) */

    return output;
}
\end{lstlisting}

In the above source-code, the first two conditional structures verify whether the \verb|Agent| has been allocated or not, and if the number of decision variables is greater than $1$. The next line implements the function itself: since $\vec{x}\in\Re^2$, each agent has two dimensions only, i.e., \verb|a->x[0]| and \verb|a->x[1]|. Notice LibOPT uses \verb|double| as the data type to allow a more accurate precision.

\begin{sloppypar}
Although the user can implement any function to be optimized, we need to follow the guidelines implemented in \verb|LibOPT/include/common.h| by the following function: \verb|typedef double (*prtFun)(Agent *, va_list arg)|. This signature tells us the function to be minimized should return a \verb|double| value, as well as its first parameter should be an \verb|Agent|, followed by a list of arguments, which depends on the function.
\end{sloppypar}

In our example, suppose we want to use Particle Swarm Optimization to minimize \verb|MyFunction|. We need first to define the parameters according to the model according to the Section~\ref{ss.model_files}. In this case, for the sake of explanation, we will use a similar model file to the one given in that section, as follows:

\begin{verbatim}
10 2 100 #<n_particles> <dimension> <max_iterations>
1.7 1.7 #<c1> <c2>
0.7 0.0 0.0 #<w> <w_min> <w_max>
-10 10 #<LB> <UB> x[0]
-10 10 #<LB> <UB> x[1]
\end{verbatim}
Notice we have only one decision variable to be optimized, as defined in the first line of the model file. Therefore, as the boundaries at the end of the file, we have set $\vec{x}\in[-10,10]$.

Let \verb|pso_model.txt| be the file name concerning the above model. Basically, one needs to create a main file to call PSO procedure as follows:

\begin{lstlisting}
#include "common.h"
#include "function.h"
#include "pso.h"

int main(){
    SearchSpace *s = NULL;
    
    s = ReadSearchSpaceFromFile("pso_model.txt", _PSO_);     
    InitializeSearchSpace(s, _PSO_);
    
    if(CheckSearchSpace(s, _PSO_)) 
        runPSO(s, MyFunction);
    
    DestroySearchSpace(&s, _PSO_); 
        
    return 0;
}
\end{lstlisting}
As one can observe, it is quite simple to execute PSO, since we need to call five main functions only:

\begin{itemize}
	\item \verb|ReadSearchSpaceFromFile|: it reads the model file and creates a search space;
	\item \verb|InitializeSearchSpace|: it initializes the search space;
	\item \verb|CheckSearchSpace|: it checks wether the search space is valid or not;
	\item \verb|runPSO|: it minimizes function \verb|MyFunction|; and
	\item \verb|DestroySearchSpace|: it deallocates the search space.
\end{itemize}
Notice one can find a number of similar examples in \verb|LibOPT/examples|.

\subsection{Adding New Techniques}
\label{ss.nre_technique}

In this section, we discuss how to add a new technique inside LibOPT~\footnote{A more detailed explanation about that topic can be found at \url{https://github.com/jppbsi/LibOPT/wiki/How-to-add-a-new-technique\%3F}}. Let us consider a fictitious optimization algorithm called Brazilian Soccer Optimization (BSO), and the following steps:

\begin{enumerate}
\begin{sloppypar}
	\item Add the following line in \verb|LibOPT/include/opt.h|: \verb|#define _BSO_ X|, where \verb|X| stands for a natural number not used before. This parameter (directive) stands for an unique number used as the metaheuristic technique identifier.
\end{sloppypar}
	\item If your technique does need a different structure not implemented in LibOPT, you must do the following:
	\begin{enumerate}
		\item In the structure \verb|Agent|, add your desired parameters. For instance, suppose BSO needs a player's strength for each decision variable. Thus, we need to consider the following structure:
	\begin{lstlisting}
typedef struct Agent_{
    int n; /* number of decision variables */  
    double *x; /* position */  
    double *v; /* velocity */  
    double f; /* fitness value */  
    ...  
    double *strength; /* >>> NEW LINE HERE <<< */
}Agent;
	\end{lstlisting}
	\item In the structure \verb|SearchSpace|, add your desired parameters. For instance, suppose BSO uses an additional variable that encodes the quality of the grass during the match: we need to add the following line:
	\begin{lstlisting}
typedef struct SearchSpace_{
    int m; /* number of agents (solutions) */
    int n; /* number of decision variables */
    Agent **a; /* array of pointers to agents */
    ...  
    double grass_quality; /* >>> NEW LINE HERE <<< */
}SearchSpace;
	\end{lstlisting}
	\item In function \verb|CreateAgent| (\verb|LibOPT/src/common.c|), you should add one more switch command in order to allocate your new variable, as well as to initialize it:
	\begin{lstlisting}
/* It creates an agent
Parameters:
n: number of decision variables
opt_id: identifier of the optimization technique */
Agent *CreateAgent(int n, int opt_id){
    if((n < 1) || opt_id < 1){
        fprintf(stderr,"\nInvalid parameters @CreateAgent.\n");
        return NULL;
    }
    	
    Agent *a = NULL;
    a = (Agent *)malloc(sizeof(Agent));
    a->v = NULL;
    a->strength = NULL; /* >>> NEW LINE HERE <<< */

    switch (opt_id){
        case _PSO_:
            a->v = (double *)malloc(n*sizeof(double));
        break;
        ...
        case _BSO_: /* >>> NEW CASE HERE <<< */
            a->strength = (double *)malloc(n*sizeof(double));
        break;
        default:
            free(a);
            fprintf(stderr,"\nInvalid optimization identifier @CreateAgent\n");
            return NULL;
        break;
    }
    
    a->x = (double *)malloc(n*sizeof(double));
    
    return a;
}	
	\end{lstlisting}
	\item In function \verb|DestroyAgent| (\verb|LibOPT/src/common.c|), you should deallocate your new variable:
	\begin{lstlisting}
/* It deallocates an agent
Parameters:
a: address of the agent to be deallocated
opt_id: identifier of the optimization technique */
void DestroyAgent(Agent **a, int opt_id){
    Agent *tmp = NULL;

    tmp = *a;
    if(!tmp){
        fprintf(stderr,"\nAgent not allocated @DestroyAgent.\n");
        exit(-1);
    }
    
    if(tmp->x) free(tmp->x);
    switch (opt_id){
        case _PSO_:
            if(tmp->v) free(tmp->v);
        break;
        case _BSO_:
            if(tmp->strength) free(tmp->strength); /* >>> DEALLOCATE YOUR VARIABLE HERE <<<*/
        break;   
        default:
            fprintf(stderr,"\nInvalid optimization identifier @DestroyAgent.\n");
        break;     
    }
    
    free(tmp);
}
	\end{lstlisting}
	\item In function \verb|CreateSearchSpace| (\verb|LibOPT/src/common.c|), you should add one more switch command in order to allocate your new variable, as well as to initialize it. Notice you must do that only if you have an array-like variable.
	\begin{lstlisting}
/* It creates a search space
Parameters:
m: number of agents
n: number of decision variables
opt_id: identifier of the optimization technique */
SearchSpace *CreateSearchSpace(int m, int n, int opt_id){
    SearchSpace *s = NULL;
    int i;
    
    if((m < 1) || (n < 1) || (opt_id < 1)){
        fprintf(stderr,"\nInvalid parameters @CreateSearchSpace.\n");
        return NULL;
    }
    
    s = (SearchSpace *)malloc(sizeof(SearchSpace));
    s->m = m;
    s->n = n;
    s->a = (Agent **)malloc(s->m*sizeof(Agent *));
    
    s->a[0] = CreateAgent(s->n, opt_id);
    if(s->a[0]){
        for(i = 1; i < s->m; i++)
            s->a[i] = CreateAgent(s->n, opt_id);
    }else{
        free(s->a);
        free(s);
        return NULL;
    }
    
    switch (opt_id){
        case _BSO_:
            /* >>> NEW VARIABLE HERE <<<*/
        break;        
    }
    
    return s;
}	
	\end{lstlisting}
	\item In function \verb|DestroySearchSpace|, you should deallocate your new variable.
	\begin{lstlisting}
/* It deallocates a search space
Parameters:
s: address of the search space to be deallocated
opt_id: identifier of the optimization technique */
void DestroySearchSpace(SearchSpace **s, int opt_id){
    SearchSpace *tmp = NULL;
    int i;
    
    tmp = *s;
    if(!tmp){
        fprintf(stderr,"\nSearch space not allocated @DestroySearchSpace.\n");
        exit(-1);
    }
    
    for(i = 0; i < tmp->m; i++)
        if(tmp->a[i]) DestroyAgent(&(tmp->a[i]), opt_id);
            free(tmp->a);
    
    switch (opt_id){
        case _BSO_:
            /* >>> DEALLOCATE YOUR VARIABLE HERE <<<*/
        break;        
        default:
            fprintf(stderr,"\nInvalid optimization identifier @DestroySearchSpace.\n");
        break;
    }
    
    free(tmp);
}	
	\end{lstlisting}
	\end{enumerate}
	\item Further, one needs to create files \verb|LibOPT/include/bso.h| and \verb|LibOPT/src/bso.c|.
	\begin{enumerate}
	\item in file \verb|LibOPT/include/bso.h|, add the following code:
	\begin{lstlisting}
#ifndef BSO_H
#define BSO_H

#include opt.h
... /* >>> YOUR NEW CODE HERE <<< */
#endif
 	\end{lstlisting}
 	\item in file \verb|LibOPT/src/bso.c|, add the following code:
	\begin{lstlisting}
#include "bso.h"	
	\end{lstlisting}
	\end{enumerate}
	\item Finally, you need to update \verb|Makefile| in order to compile your new technique. You can just copy and paste the lines regarding any technique that has been written already.
\end{enumerate} 

\section{Conclusions}
\label{s.conclusions}

In this paper, we presented an open-source library for handling metaheuristic techniques and function optimization called LibOPT. The main features of the library concerns fast prototyping, self-contained code, as well as a simple but efficient implementation.

The library implements a number of optimization techniques, as well as more than a hundred of benchmark functions. Additionally, LibOPT implements hypercomplex-based search spaces, which we believe makes it one the first of its kind in the literature. We also showed how to use LibOPT to optimize functions, as well as how to add  your own technique. Currently, LibOPT's implementation is for nonlinear optimization problems with simple bounds.

In regard to future works, we intend to make available multi- and many-objective versions of the techniques, to support constraint-handling techniques, such as penalty method, to extend its usage to discrete problems, as well as to support more efficient implementations based on Graphics Processing Units.

\begin{acks}
The authors are grateful to FAPESP grants \#2013/20387-7, \#2014/12236-1, \#2014/16250-9, \#2015/25739-4, and \#2016/21243-7, CNPq grant \#306166/2014-3, and Capes.
\end{acks}

\bibliographystyle{ACM-Reference-Format}
\bibliography{references} 

\end{document}